\documentclass[runningheads,a4paper]{llncs}

\usepackage{amsmath}
\usepackage{amssymb}
\usepackage{graphicx}
\usepackage{tabularx}
\usepackage{booktabs}
\usepackage{microtype}
\usepackage{tikz,pgfplots}
\usepackage{tikz-3dplot}
\usepackage{algorithm}
\usepackage{algpseudocode}
\usetikzlibrary{spy}

\usepackage{hyperref}

\newcommand{\eg}{e.\,g.\,\,}

\newcommand{\wrt}{w.r.t.\,\,}
\def\argmin{\mathop{\mathrm{argmin}}}

\newlength \figureheight
\newlength \figurewidth
\usepackage{subfig}

\begin{document}

\mainmatter  


\title{QuaSI: Quantile Sparse Image Prior for Spatio-Temporal Denoising of Retinal OCT Data}

\authorrunning{Schirrmacher, K\"ohler et al.}
\titlerunning{QuaSI: Quantile Sparse Image Prior for Spatio-Temporal Denoising}

%
%
\author{Franziska~Schirrmacher$^{1,*}$, Thomas~K\"ohler$^{1,*}$, Lennart~Husvogt$^{1}$, James~G.~Fujimoto$^{2}$, Joachim~Hornegger$^{1}$, and~Andreas~K.~Maier$^{1}$}
%

\institute{$^{1}$~Pattern Recognition Lab, Friedrich-Alexander-Universit{\"a}t Erlangen-N{\"u}rnberg\\
\email{\{franziska.schirrmacher, thomas.koehler\}@fau.de}\\
$^{2}$~Department of Electrical Engineering \& Computer Science and Research Laboratory of Electronics, Massachusetts Institute of Technology \\[0.2cm]
$^*$ These authors contributed equally to this work.
}

%
%

\maketitle

\begin{abstract}
Optical coherence tomography (OCT) enables high-resolution and non-invasive 3D imaging of the human retina but is inherently impaired by speckle noise. This paper introduces a spatio-temporal denoising algorithm for OCT data on a B-scan level using a novel quantile sparse image (QuaSI) prior. To remove speckle noise while preserving image structures of diagnostic relevance, we implement our QuaSI prior via median filter regularization coupled with a Huber data fidelity model in a variational approach. For efficient energy minimization, we develop an alternating direction method of multipliers (ADMM) scheme using a linearization of median filtering. Our spatio-temporal method can handle both, denoising of single B-scans and temporally consecutive B-scans, to gain volumetric OCT data with enhanced signal-to-noise ratio. Our algorithm based on 4 B-scans only achieved comparable performance to averaging 13 B-scans and outperformed other current denoising methods.
\end{abstract}

\section{Introduction}

Since its invention in 1991, optical coherence tomography (OCT) \cite{Choi2013a} has become a standard imaging technique within clinical workflows in ophthalmology. OCT enables non-invasive 3D imaging of retinal layers with spatial resolutions in a micrometer range. These properties contributed to its wide distribution for diagnosis and disease monitoring as well as in novel systems for computer-aided diagnostics \cite{Kohler2015}. Beside these merits, OCT suffers from a low signal-to-noise ratio (SNR) due to speckle noise caused by photon interference. Hardware-based techniques, e.\,g. frequency compounding \cite{Pircher2003}, are able to reduce speckle noise but increase the complexity of instruments or scanning protocols. For this reason, image-based post-processing algorithms to improve the reliability of retinal OCT data are attractive. However, many popular denoising methods such as BM3D \cite{Dabov2007} have been mainly designed for natural images and can handle non-Gaussian noise only to a certain extend. Furthermore, noise reduction in OCT is a sensitive issue as the preservation of tiny morphological structures is a mandatory requirement. \textit{Spatial} or \textit{single-image} methods perform noise reduction by filtering single B-scans. Spatial filters suited for OCT are hybrid median filter, Lee filter, Wiener filter, or wavelet thresholding \cite{Ozcan2007}. Popular global methods include non-linear diffusion \cite{Salinas2007}, variational formulations \cite{Duan2016}, or structure-adaptive Bayesian estimation \cite{Wong2010}. More recently, sparse coding \cite{Fang2012} using high SNR scans for dictionary learning to denoise single low SNR scans has been proposed. The spatial methods have in common that noise reduction is limited as they utilize single B-scans only. \textit{Temporal} or \textit{multi-image} methods exploit several B-scans acquired sequentially from the same location or nearby positions. A simple method often implemented for commercial systems is averaging of multiple registered B-scans. More recent approaches are wavelet multi-image denoising \cite{Mayer2012} or matrix completion \cite{Cheng2014} that have been customized to speckle noise reduction and outperform simple averaging. However, temporal methods require longer acquisition times to gain multiple B-scans and hence increase patient discomfort. 

This paper introduces a new \textit{spatio-temporal} OCT denoising algorithm. Our contribution is two-fold: 1) We propose denoising via energy minimization based on the novel class of quantile sparse image (QuaSI) priors. To deal with speckle noise while preserving morphological structures, we regularize with the popular median filter as a special instance of our QuaSI prior. 2) We develop an alternating direction method of multipliers (ADMM) scheme for optimization with the non-linear median filter. The proposed spatio-temporal method can handle both, denoising of single and multiple registered B-scans. 

\section{Proposed Spatio-Temporal Denoising Algorithm}
\label{sec:ProposedSpatioTemporalDenoisingAlgorithm}

\subsection{Noise Model and Energy Minimization Formulation}
\label{sec:NoiseModelAndEnergyMinimizationFormulation}

Our method aims at denoising volumetric OCT data, where a single volume is represented as a stack of B-scans $\tilde{\vec{G}} \in \mathbb{R}^{L \times N_x \times N_y}$. We denote the $l$-th B-scan of size $N_x \times N_y$ in vector notation as $\tilde{\vec{g}}_l \in \mathbb{R}^{N}$ with $l \in [1, L]$ and $N = N_x N_y$. Each noisy B-scan $\tilde{\vec{g}}_l$ in a given volume is related to the respective noise-free scan $\tilde{\vec{f}}_l$ according to multiplicative speckle noise. Following related denoising methods \cite{Wong2010,Duan2016}, we model speckle noise in a logarithmic measurement range according to $\vec{f}_l = \vec{g}_l + \vec{n}_l$, where $\vec{f}_l = \log(\tilde{\vec{f}}_l)$, $\vec{g} = \log(\tilde{\vec{g}}_l)$ and $\vec{n}_l \in \mathbb{R}^{N}$ is additive noise. 

Let $\vec{g}^{(1)}, \ldots, \vec{g}^{(K)}$ be a set of B-scans that are captured from the same location and registered to each other. We estimate a denoised B-scan $\hat{\vec{f}}$ according to: 
\begin{equation}
	\hat{\vec{f}} = \argmin_{\vec{f}} \mathcal{L}(\vec{f}) 
	= \argmin_{\vec{f}} \sum_{k = 1}^{K}  \rho \big( \vec{f} - \vec{g}^{(k)} \big) + \mu \Vert\nabla \vec{f}\Vert_{1} + \lambda R_{\mathrm{QuaSI}}(\vec{f}).
	\label{eqn:objective}
\end{equation}
In \eqref{eqn:objective}, the first term denotes the fidelity of $\vec{f}$ \wrt $K$ observed noisy B-scans $\vec{g}^{(k)}$, $k = 1, \ldots, K$ according to the loss function $\rho: \mathbb{R}^N \rightarrow \mathbb{R}_0^+$. The second term is the anisotropic total variation (TV) to regularize the image gradient $\nabla\vec{f} = (\nabla_{x}\vec{f}, \nabla_{y}\vec{f})^{\top}$ with the weight $\mu \geq 0$. The third term denotes regularization according to our proposed QuaSI prior with the weight $\lambda \geq 0$, see Section \ref{sec:QuantileSparseImagePrior}. 

To define the data fidelity term in \eqref{eqn:objective}, we propose to use outlier-insensitive loss functions. In this paper, we use the Huber loss \cite{Ochs2015} for $\rho(\cdot)$ to model the image formation in retinal OCT. It is worth noting that the Huber loss can tackle outliers related to non-Gaussian noise, motion artifacts, or misregistrations of consecutive B-scans while being convex and easy to optimize.

\subsection{Quantile Sparse Image Prior}
\label{sec:QuantileSparseImagePrior}

The novel class of priors that we propose is based on quantile filtering. We denote a quantile filter as $\tilde{\vec{f}} = Q(\vec{f})$, where $\tilde{f}_i = \mathrm{quantile}_{\mathcal{N}(i)} (f_i, p)$ determines the $p$-quantile with $p \in [0, 1]$ within the local neighborhood $\mathcal{N}(i)$ centered at the $i$-th pixel in $\vec{f}$. Our prior model is defined as fixed point under the quantile filter according to:
\begin{equation}
	R_{\mathrm{QuaSI}}(\vec{f}) = \left|\left| \vec{f} - \textit{Q}(\vec{f}) \right|\right|_1.
	\label{eqn:quasiPrior}
\end{equation}
Similar forms have also become popular in other inverse problems, \eg the dark channel prior \cite{Pan2016} or regularization by denoising priors \cite{Romano2016a}, and our model is inspired by such concepts. This general model facilitates regularization by various types of order statistics or parameters such as minimum or maximum, first or third quartile, weighted median, etc. To customize \eqref{eqn:quasiPrior} for OCT denoising in this paper, we use the $p = 0.5$ quantile that is equivalent to the median. This implements a regularization by the popular median filter within our spatio-temporal framework. We found that this facilitates structure-preserving denoising and handles non-Gaussian noise as important prerequisites for the desired application.

\subsection{Alternating Direction Method of Multiplier Optimization}
\label{sec:AlternatingDirectionMethodOfMultiplierADMMOptimization}

The optimization of \eqref{eqn:objective} involves two non-smooth regularization terms related the TV prior and our proposed QuaSI prior. For efficient minimization with the non-smooth terms, we adopt ADMM optimization \cite{Goldstein2009}. To this end, we introduce the auxiliary variables $\vec{u}$ and $\vec{v}$ and re-formulate \eqref{eqn:objective} via the augmented Lagrangian:
\begin{equation}
	\begin{split}
		\mathcal{L}_{\mathrm{AL}}(\vec{f}, \vec{u}, \vec{v}, \vec{b}_u, \vec{b}_v) 
		= \sum_{k = 1}^{K} & \rho \big( \vec{f} - \vec{g}^{(k)} \big) + \dfrac{\alpha}{2} \Vert \vec{u} - \vec{f} + \textit{Q}(\vec{f}) - \vec{b}_{u} \Vert_{2}^{2} \\ 
		&+ \lambda\Vert \vec{u} \Vert_{1} + \dfrac{\beta}{2} \Vert \vec{v} - \nabla \vec{f} - \vec{b}_v\Vert_{2}^{2} + \mu\Vert \vec{v} \Vert_{1},
	\end{split}
	\label{eqn:augmentedLagrangian}
\end{equation}
where $\vec{b}_{u}$ and $\vec{b}_{v}$ denote Bregman variables, and $\alpha > 0$ and $\beta > 0$ are Lagrangian multiplier to enforce the constraints $\vec{u} = \textit{Q}(\vec{f})$ and $\vec{v} = \nabla \vec{f}$. We iteratively optimize \eqref{eqn:augmentedLagrangian} by alternating minimization \wrt the individual parameters. 

Notice that direct optimization of \eqref{eqn:augmentedLagrangian} is not tractable due to the non-linearity of the quantile operator $\textit{Q}(\vec{f})$ and the quantile operator is first linearized as $\textit{Q}(\vec{f}) = \vec{Q} \vec{f}$ \cite{Pan2016}. If $\textit{Q}(\vec{f})$ denotes median filtering, we assemble $\vec{Q}$ element-wise as a binary matrix according to $Q_{ij} = 1 \Leftrightarrow z = j$, where $j = \arg\mathrm{median}_{z \in \mathcal{N}(i)} f_z$ denotes the position of the median in the neighborhood $\mathcal{N}(i)$ centered at the $i$-th pixel. This construction fulfills $\textit{Q}(\vec{f}^\prime) = \vec{Q} \vec{f}^\prime$ for $\vec{f}^\prime = \vec{f}$, while otherwise we use $\vec{Q}$ as an approximation of the median filter. In the proposed ADMM scheme, we gradually update $\vec{Q}$ and assemble the linearization from the intermediate image $\vec{f}^t$ estimated at the previous iteration. Given this linearization, we minimize \eqref{eqn:augmentedLagrangian} \wrt the denoised image $\vec{f}$. To handle the Huber loss, this is done by iteratively re-weighted least squares (IRLS). This leads to the linear system:
\begin{equation}
	\begin{split}
		\bigg[ & 2 \sum_{k=1}^K \vec{W}^{(k)} + \alpha \left(\vec{I} - \vec{Q} \right)^\top \left(\vec{I} - \vec{Q} \right) + \beta \nabla^\top \nabla \bigg] \vec{f}^{t + 1} \\
		= ~ & 2  \sum_{k = 1}^K \vec{W}^{(k)} \vec{g}^{(k)} + \alpha \left(\vec{I} - \vec{Q}\right)^\top (\vec{u} - \vec{b}_u) + \beta \nabla^\top (\vec{v} - \vec{b}_v),
	\end{split}
	\label{eqn:cgEquationSystem}
\end{equation}
where $\vec{W}^{(k)}$ denotes a diagonal weight matrix derived from the Huber loss. For IRLS, the weights are computed based on the intermediate image $\vec{f}^t$ according to $W_{ii}^{(k)} = \rho^\prime(f_i - g_i^{(k)}) / (f^t_i - g_i^{(k)})$, where $\rho^\prime(z)$ is the derivative of the Huber loss \cite{Ochs2015}. The linear system in \eqref{eqn:cgEquationSystem} is solved by conjugate gradient (CG) iterations.

The minimization of \eqref{eqn:augmentedLagrangian} \wrt to the auxiliary variables is separable and performed element-wise. We obtain closed-form solutions using shrinkage operations:
\begin{align}
	u_i^{t+1} &= \mathrm{shrink} \left( [\vec{f}^{t+1} - \vec{Q} \vec{f}^{t+1} + \vec{b}^{t}_{u}]_i, \lambda/\alpha \right), \\
	v_i^{t+1} &= \mathrm{shrink} \left( [\nabla \vec{f}^{t+1} + \vec{b}^{t}_v]_i, \mu / \beta \right),  
	\label{eqn:updateAux}
\end{align}
where $\mathrm{shrink}(z, \gamma) = \mathrm{sign}(z) \max(z - \gamma, 0)$ denotes soft-thresholding associated with the $L_1$ norm \cite{Goldstein2009}. Finally, the Bregman variables are updated according to:
\begin{align}
	\vec{b}^{t+1}_u &= \vec{b}^{t}_u + (\vec{f}^{t+1} - \vec{Q} \vec{f}^{t+1} - \vec{u}^{t+1}), \\
	\vec{b}^{t+1}_v &= \vec{b}_v^{t} + (\nabla\vec{f}^{t+1} - \vec{v}^{t+1}).
	\label{eqn:updateBregman}
\end{align}

Algorithm \ref{alg:denoisingAlgorithm} summarizes ADMM using \smash{$\vec{u}^1 = \vec{v}^1 = \vec{0}$, $\vec{b}_u^1 = \vec{b}_v^1 = \vec{0}$}, and the average of the input B-scans as an initial guess \smash{$\vec{f}^1$}. We found that for convergence it is sufficient to update the median filter linearization only after a couple of iterations to speed up ADMM. For this reason, we use $T_{\mathrm{inner}}$ iterations to update $\vec{f}, \vec{u}$, $\vec{v}$, $\vec{b}_u$, and $\vec{b}_v$ and $T_{\mathrm{outer}}$ iterations to update $\vec{Q}$.
\begin{algorithm}[!t]
	\caption{Spatio-temporal denoising algorithm using ADMM optimization}
	\label{alg:denoisingAlgorithm}
	\begin{algorithmic}
		\State Initialize $\vec{u}^1 = \vec{v}^1 = \vec{0}$, $\vec{b}_u^1 = \vec{b}_v^1 = \vec{0}$ and $\vec{f}^1 = \mathrm{mean}(\vec{g}^{(1)}, \ldots, \vec{g}^{(K)})$
		\For{$t = 1, \ldots, T_{\mathrm{outer}}$}
			\State Assemble $\vec{Q}$ from the intermediate image $\vec{f}^t$
			\For{$i = 1, \ldots, T_{\mathrm{inner}}$}
				\State Update the intermediate image $\vec{f}^{t+1}$ using CG iterations for \eqref{eqn:cgEquationSystem}
				\State Update the auxiliary variables $\vec{u}^{t+1}$ and $\vec{v}^{t+1}$ using \eqref{eqn:updateAux}
				\State Update the Bregman variables $\vec{b}_u^{t+1}$ and $\vec{b}_v^{t+1}$ using \eqref{eqn:updateBregman}
			\EndFor
		\EndFor
	\end{algorithmic}
\end{algorithm}

\section{Experiments and Results}
\label{sec:ExperimentsAndResults}

\begin{figure}[!t]
 \scriptsize
 \centering
 \subfloat{\includegraphics[width = 0.60\textwidth]{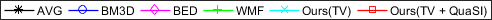}}\\[-0.8em]
 \setlength \figurewidth{0.41\textwidth}
 \setlength \figureheight{0.60\figurewidth}
 \subfloat{
%
%
\definecolor{mycolor1}{rgb}{1.00000,0.00000,1.00000}%
\definecolor{mycolor2}{rgb}{0.00000,1.00000,1.00000}%
\begin{tikzpicture}

\begin{axis}[%
width=0.951\figurewidth,
height=\figureheight,
at={(0\figurewidth,0\figureheight)},
scale only axis,
unbounded coords=jump,
xmin=1,
xmax=13,
xlabel style={font=\color{white!15!black}},
xlabel={Number of input images},
ymin=18,
ymax=29,
ylabel style={font=\color{white!15!black}},
ylabel={PSNR value},
axis background/.style={fill=white},
xmajorgrids,
ymajorgrids,
ylabel near ticks,
xlabel near ticks
]
\addplot [color=black, line width=1.0pt, mark size = 1.5pt, mark=asterisk, mark options={solid, black}, forget plot]
  table[row sep=crcr]{%
1	16.5810358243587\\
2	19.2729662352964\\
3	20.756070498541\\
4	21.7169287969579\\
5	22.4477242696223\\
6	23.0119572550341\\
7	23.4794721345791\\
8	23.8626334425206\\
9	24.1734681746898\\
10	24.4362922354348\\
11	24.6786113415413\\
12	24.9077115471953\\
13	25.1334021829532\\
};
\addplot [color=blue, line width=1.0pt, mark size = 1.5pt, mark=o, mark options={solid, blue}, forget plot]
  table[row sep=crcr]{%
1	18.3224832361918\\
2	21.9288895613494\\
3	23.9311610992459\\
4	25.1320243506136\\
5	25.9963652654202\\
6	26.6254529168428\\
7	27.1062280669924\\
8	27.4942861484507\\
9	27.7845590412387\\
10	28.006203190586\\
11	28.2200443364765\\
12	28.4201885652835\\
13	28.6705348247397\\
};
\addplot [color=mycolor1, line width=1.0pt, mark size = 1.5pt, mark=diamond, mark options={solid, mycolor1}, forget plot]
  table[row sep=crcr]{%
1	21.3174876429947\\
2	23.3615864314527\\
3	24.5278978001372\\
4	25.208538924165\\
5	25.7205373177648\\
6	26.1103734193355\\
7	26.4232124462734\\
8	26.6983759041042\\
9	26.900388615591\\
10	27.0604320635477\\
11	27.2204895246033\\
12	27.3848673339503\\
13	27.5905596452095\\
};
\addplot [color=green, line width=1.0pt, mark size = 1.5pt, mark=+, mark options={solid, green}, forget plot]
  table[row sep=crcr]{%
1	nan\\
2	23.0841457979936\\
3	24.6766826132431\\
4	25.3948452744788\\
5	26.0745419019233\\
6	26.5152051787939\\
7	26.9197368664109\\
8	27.2188144821966\\
9	27.473397639023\\
10	27.6545370201535\\
11	27.8549623811908\\
12	28.031939294805\\
13	28.2669117681537\\
};
\addplot [color=mycolor2, line width=1.0pt, mark size = 1.5pt, mark=x, mark options={solid, mycolor2}, forget plot]
  table[row sep=crcr]{%
1	19.7703753113614\\
2	22.7068417630753\\
3	24.3443476068201\\
4	25.4061069727842\\
5	26.1957132860467\\
6	26.7864248073808\\
7	27.2310435978842\\
8	27.5993607698691\\
9	27.8542692964733\\
10	28.0462828166795\\
11	28.2436610420909\\
12	28.4204558498466\\
13	28.6672584022061\\
};
\addplot [color=red, line width=1.0pt, mark size = 1.5pt, mark=square, mark options={solid, red}, forget plot]
  table[row sep=crcr]{%
1	20.6567924530505\\
2	23.4123924667701\\
3	25.009569260864\\
4	25.9840023556481\\
5	26.7014290527431\\
6	27.23041751942\\
7	27.6230227833808\\
8	27.9577472536867\\
9	28.1789691141212\\
10	28.3386123314719\\
11	28.514418340608\\
12	28.6725716180855\\
13	28.9132400923243\\
};
\end{axis}
\end{tikzpicture}%
}\qquad
 \subfloat{
%
%
\definecolor{mycolor1}{rgb}{1.00000,0.00000,1.00000}%
\definecolor{mycolor2}{rgb}{0.00000,1.00000,1.00000}%
\begin{tikzpicture}

\begin{axis}[%
width=0.951\figurewidth,
height=\figureheight,
at={(0\figurewidth,0\figureheight)},
scale only axis,
unbounded coords=jump,
xmin=1,
xmax=13,
xlabel style={font=\color{white!15!black}},
xlabel={Number of input images},
ymin=0.2,
ymax=0.9,
ylabel style={font=\color{white!15!black}},
ylabel={SSIM value},
axis background/.style={fill=white},
xmajorgrids,
ymajorgrids,
ylabel near ticks,
xlabel near ticks
]
\addplot [color=black, line width=1.0pt, mark size = 1.5pt, mark=asterisk, mark options={solid, black}, forget plot]
  table[row sep=crcr]{%
1	0.0962500609976868\\
2	0.171490714339526\\
3	0.228465176209863\\
4	0.27436971226936\\
5	0.313728460236347\\
6	0.346678374505251\\
7	0.376310260606632\\
8	0.402055522363285\\
9	0.424773840381719\\
10	0.445095090536195\\
11	0.46348739370096\\
12	0.480590497756614\\
13	0.495795391786353\\
};
\addplot [color=blue, line width=1.0pt, mark size = 1.5pt, mark=o, mark options={solid, blue}, forget plot]
  table[row sep=crcr]{%
1	0.204072921328526\\
2	0.392195279169313\\
3	0.517418238611999\\
4	0.602335454386478\\
5	0.663689672529456\\
6	0.707798798489967\\
7	0.741495276987715\\
8	0.767208139095683\\
9	0.787175687362134\\
10	0.802899390525735\\
11	0.815738120688469\\
12	0.826601197139287\\
13	0.83565672315518\\
};
\addplot [color=mycolor1, line width=1.0pt, mark size = 1.5pt, mark=diamond, mark options={solid, mycolor1}, forget plot]
  table[row sep=crcr]{%
1	0.319379224188558\\
2	0.436022053744835\\
3	0.503588417991245\\
4	0.549338561742467\\
5	0.583921610812764\\
6	0.610426469784646\\
7	0.6323217865256\\
8	0.650695761572665\\
9	0.665995289806721\\
10	0.678975127155537\\
11	0.690526172283753\\
12	0.70123857839236\\
13	0.710766093856486\\
};
\addplot [color=green, line width=1.0pt, mark size = 1.5pt, mark=+, mark options={solid, green}, forget plot]
  table[row sep=crcr]{%
1	nan\\
2	0.424287388894046\\
3	0.514215637140426\\
4	0.565349007496408\\
5	0.614552608581872\\
6	0.647179620850339\\
7	0.67853782763411\\
8	0.700980805791096\\
9	0.722027709823834\\
10	0.738143992833489\\
11	0.753567565108423\\
12	0.766027434032479\\
13	0.777684410429611\\
};
\addplot [color=mycolor2, line width=1.0pt, mark size = 1.5pt, mark=x, mark options={solid, mycolor2}, forget plot]
  table[row sep=crcr]{%
1	0.26728902474242\\
2	0.459799895293525\\
3	0.576034570869754\\
4	0.655868174747303\\
5	0.712008339237269\\
6	0.751637806677059\\
7	0.780901047426126\\
8	0.802695936387726\\
9	0.818712163302508\\
10	0.831320135427656\\
11	0.84133989009626\\
12	0.84973876472427\\
13	0.856429000190367\\
};
\addplot [color=red, line width=1.0pt, mark size = 1.5pt, mark=square, mark options={solid, red}, forget plot]
  table[row sep=crcr]{%
1	0.332469828202898\\
2	0.527151701396217\\
3	0.636442089178376\\
4	0.705854804212316\\
5	0.75292429147671\\
6	0.785133352921367\\
7	0.808366946148027\\
8	0.825824638082103\\
9	0.838227236918566\\
10	0.847860297618544\\
11	0.855652122968618\\
12	0.862253500607226\\
13	0.86752368265657\\
};
\end{axis}
\end{tikzpicture}%
}
 \caption{Mean PSNR and SSIM of different denoising methods on the pig eye dataset for different numbers of input images.}
 \label{fig:pigEyeQualityMeasures}
\end{figure}
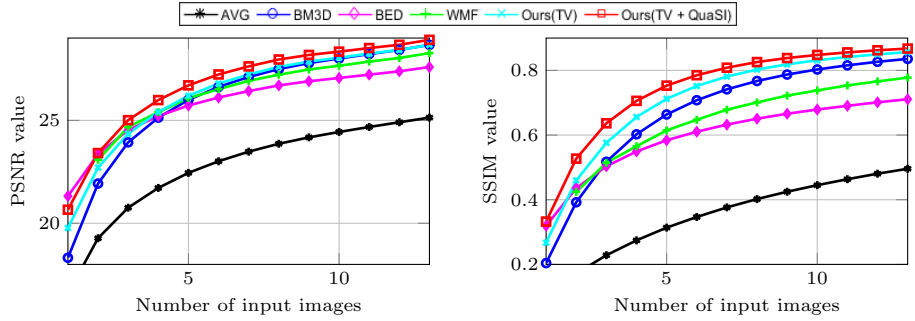

\begin{figure}[!t]
	\centering
	\scriptsize
	\subfloat[Noisy image (MSR: 2.68, CNR: 2.47)]{
		\begin{tikzpicture}[spy using outlines={rectangle,orange,magnification=2.8, height=2.7cm, width = 1.5cm, connect spies, every spy on node/.append style={thick}}]
			\node {\pgfimage[width = 0.35\textwidth]{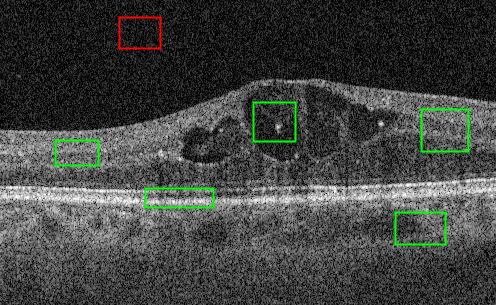}};
			\spy on (1.22,0.18) in node [left] at (3.7, 0);
		\end{tikzpicture}\label{fig:patientDataImages:noisy}}  
	\subfloat[BM3D \cite{Dabov2007} (MSR: 4.61, CNR: 4.85)]{
		\begin{tikzpicture}[spy using outlines={rectangle,orange,magnification=2.8, height=2.7cm, width = 1.5cm, connect spies, every spy on node/.append style={thick}}]
			\node {\pgfimage[width = 0.35\textwidth]{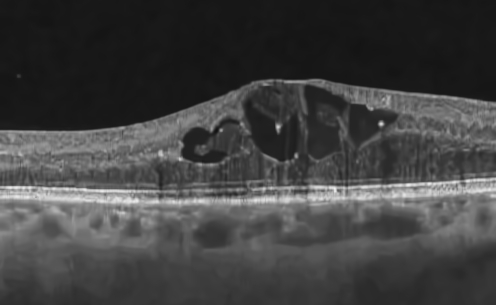}};
			\spy on (1.22,0.18) in node [left] at (3.7, 0);
    \end{tikzpicture}\label{fig:patientDataImages:bed}}\\[-1.0em]
	\subfloat[WMF \cite{Mayer2012} (MSR: 3.67, CNR: 3.55)]{
		\begin{tikzpicture}[spy using outlines={rectangle,orange,magnification=2.8, height=2.7cm, width = 1.5cm, connect spies, every spy on node/.append style={thick}}]
			\node {\pgfimage[width = 0.35\textwidth]{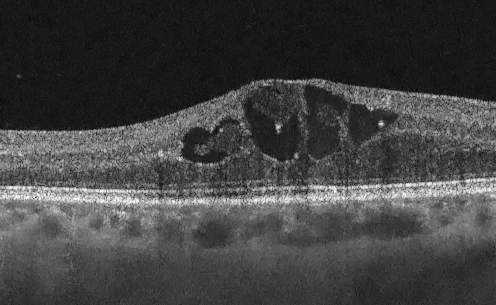}};
			\spy on (1.22,0.18) in node [left] at (3.7, 0);
    \end{tikzpicture}\label{fig:patientDataImages:wmf}} 
	\subfloat[Ours (MSR: 5.02, CNR: 5.36)]{
		\begin{tikzpicture}[spy using outlines={rectangle,orange,magnification=2.8, height=2.7cm, width = 1.5cm, connect spies, every spy on node/.append style={thick}}]
			\node {\pgfimage[width = 0.35\textwidth]{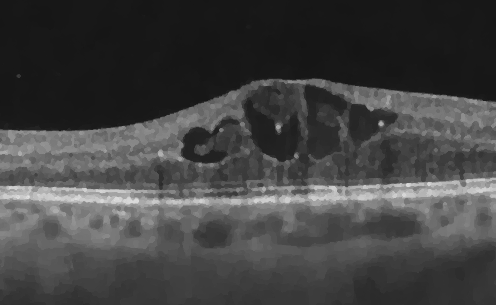}};
			\spy on (1.22,0.18) in node [left] at (3.7, 0);
    \end{tikzpicture}\label{fig:patientDataImages:ours}} 
	\caption{Denoising on our clinical dataset using $K = 5$ B-scans from a 46 years old male diabetic retinopathy patient. \protect\subref{fig:patientDataImages:noisy} Noisy image with manually selected background (red) and foreground regions (green) to determine MSR and CNR. \protect\subref{fig:patientDataImages:bed} - \protect\subref{fig:patientDataImages:ours} BED \cite{Wong2010}, WMF \cite{Mayer2012} and our proposed method.}
	\label{fig:patientDataImages}
\end{figure}

We compare our method to the well known BM3D \cite{Dabov2007} as well as current OCT noise reduction algorithms, namely Bayesian estimation denoising (BED) \cite{Wong2010}, averaging (AVG) of registered B-scans, and wavelet multi-frame denoising (WMF) \cite{Mayer2012}. As BM3D and BED handle single B-scans only, we apply these methods to the outcome of AVG for fair comparisons. WMF is a pure temporal approach and requires at least two registered B-scans. The parameters of our method were set to $\mu = 0.075 \cdot K$, $\lambda = 5.0 \cdot K$, $\alpha = 100.0 \cdot K$, $\beta = 1.5 \cdot K$, $T_{\mathrm{outer}} = 20$ and $T_{\mathrm{inner}} = 2$ for $K$ B-scans and $3 \times 3$ median filtering to setup the QuaSI prior. 
\\
\\
\textbf{Pig Eye Data.}
To study the behavior of our algorithm quantitatively, we conducted experiments on the publicly available pig eye dataset provided by Mayer et\,al. \cite{Mayer2012}. The dataset was captured ex-vivo by scanning a pig eye with a Spectralis HRA \& OCT and comprises subsets of 35 eye positions with 13 B-scans each. Following \cite{Mayer2012}, a gold standard B-scan was computed by averaging the 455 scans that have already been registered to each other. We applied the competing denoising methods on subsets of $K$ registered B-scans with $K \in [1, 13]$. 
\begin{figure}[!t]
	\centering
	\scriptsize
		\subfloat{
		\begin{tikzpicture}[spy using outlines={rectangle,orange,magnification=2.5, height=1.5cm, width = 1.5cm, connect spies, every spy on node/.append style={thick}}]
			\node {\pgfimage[width = 0.31\textwidth]{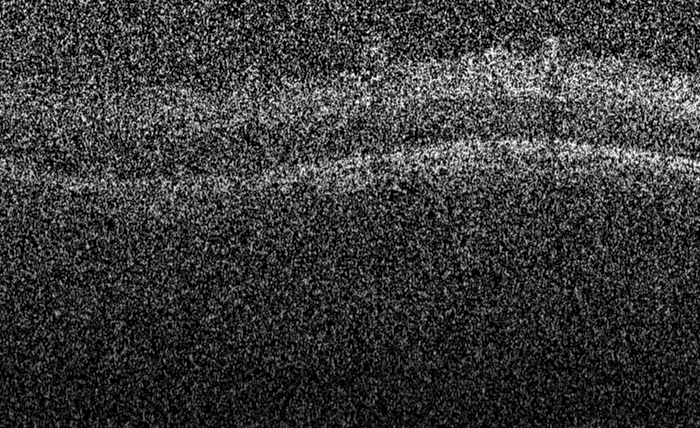}};
			\spy on (0.1,0.85) in node [left] at (1.9, -0.45);     
		\end{tikzpicture}} 
	\subfloat{
		\begin{tikzpicture}[spy using outlines={rectangle,orange,magnification=2.5, height=1.5cm, width = 1.5cm, connect spies, every spy on node/.append style={thick}}]
			\node {\pgfimage[width = 0.31\textwidth]{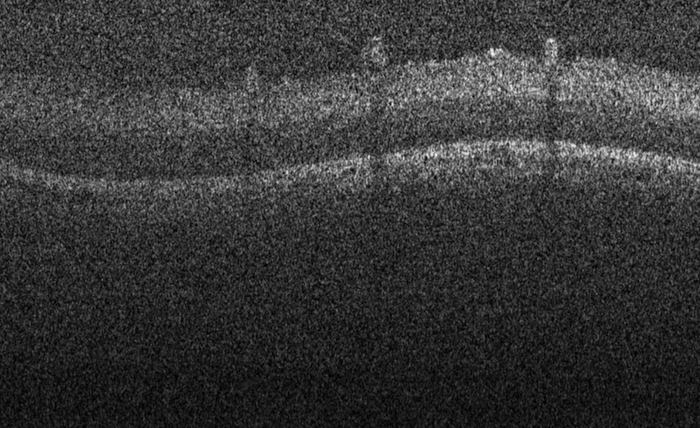}};
			\spy on (0.1,0.85) in node [left] at (1.9, -0.45);     
    \end{tikzpicture}}
	\subfloat{
		\begin{tikzpicture}[spy using outlines={rectangle,orange,magnification=2.5, height=1.5cm, width = 1.5cm, connect spies, every spy on node/.append style={thick}}]
			\node {\pgfimage[width = 0.31\textwidth]{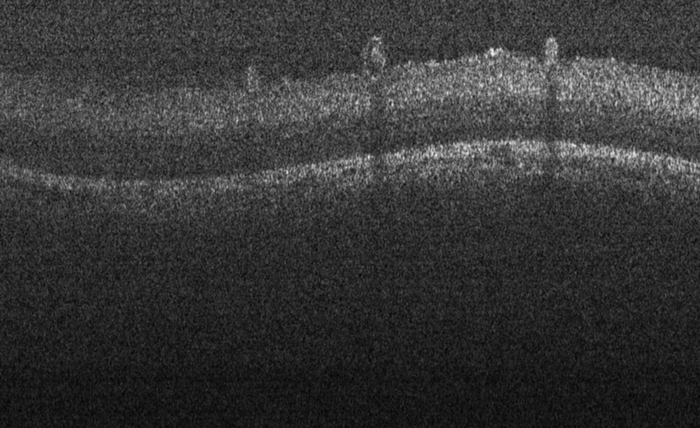}};
			\spy on (0.1,0.85) in node [left] at (1.9, -0.45);     
		\end{tikzpicture}}\\[-1.2em]
	\setcounter{subfigure}{0}
	\subfloat{
		\begin{tikzpicture}[spy using outlines={rectangle,orange,magnification=2.5, height=1.5cm, width = 1.5cm, connect spies, every spy on node/.append style={thick}}]
			\node {\pgfimage[width = 0.31\textwidth]{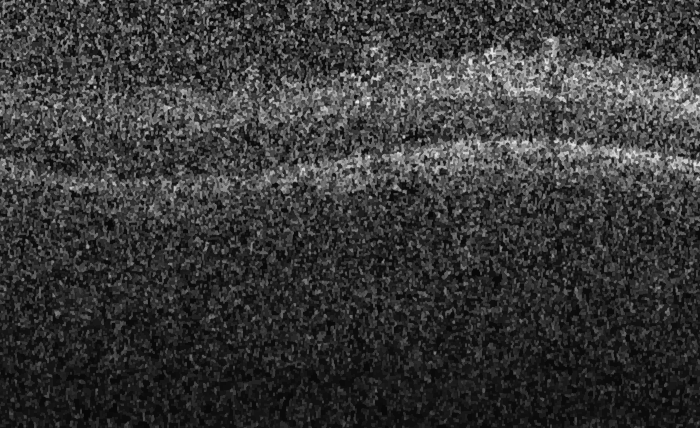}};
			\spy on (0.1,0.85) in node [left] at (1.9, -0.45);     
		\end{tikzpicture}} 
	\subfloat{
		\begin{tikzpicture}[spy using outlines={rectangle,orange,magnification=2.5, height=1.5cm, width = 1.5cm, connect spies, every spy on node/.append style={thick}}]
			\node {\pgfimage[width = 0.31\textwidth]{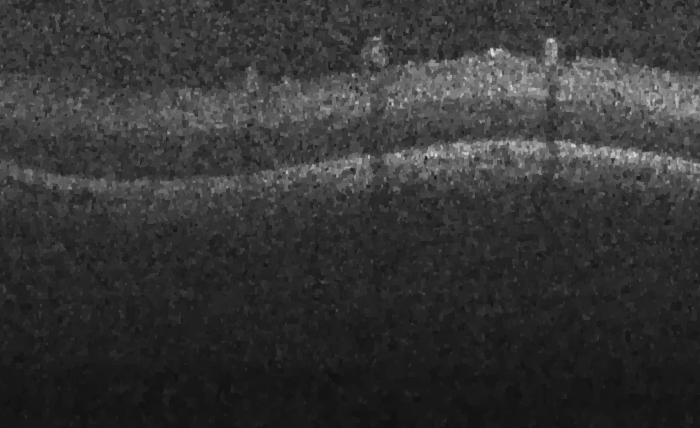}};
			\spy on (0.1,0.85) in node [left] at (1.9, -0.45);     
    \end{tikzpicture}}
	\subfloat{
		\begin{tikzpicture}[spy using outlines={rectangle,orange,magnification=2.5, height=1.5cm, width = 1.5cm, connect spies, every spy on node/.append style={thick}}]
			\node {\pgfimage[width = 0.31\textwidth]{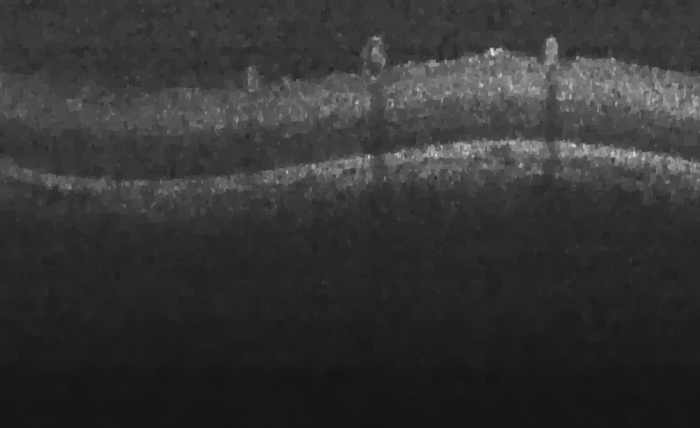}};
			\spy on (0.1,0.85) in node [left] at (1.9, -0.45);     
		\end{tikzpicture}}\\[-1.2em]
	\setcounter{subfigure}{0}
	\subfloat[$K = 1$]{
		\begin{tikzpicture}[spy using outlines={rectangle,orange,magnification=2.5, height=1.5cm, width = 1.5cm, connect spies, every spy on node/.append style={thick}}]
			\node {\pgfimage[width = 0.31\textwidth]{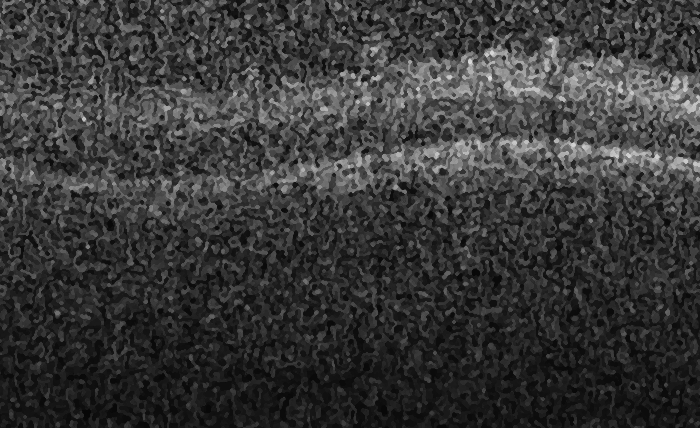}};
			\spy on (0.1,0.85) in node [left] at (1.9, -0.45);     
     \end{tikzpicture}}
		\subfloat[$K = 5$]{
			\begin{tikzpicture}[spy using outlines={rectangle,orange,magnification=2.5, height=1.5cm, width = 1.5cm, connect spies, every spy on node/.append style={thick}}]
				\node {\pgfimage[width = 0.31\textwidth]{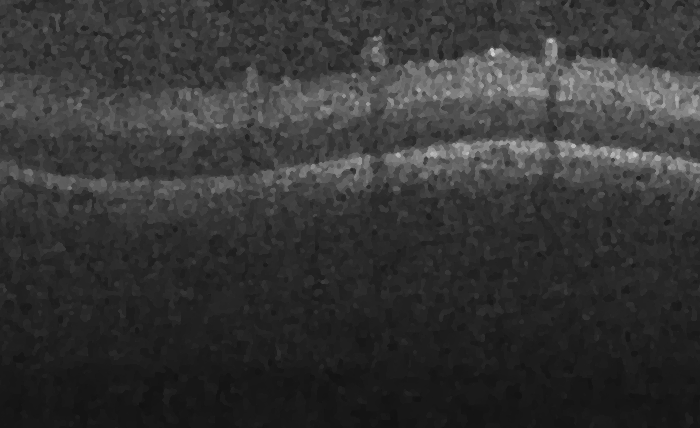}};
				\spy on (0.1,0.85) in node [left] at (1.9, -0.45);     
			\end{tikzpicture}}
	\subfloat[$K = 13$]{
		\begin{tikzpicture}[spy using outlines={rectangle,orange,magnification=2.5, height=1.5cm, width = 1.5cm, connect spies, every spy on node/.append style={thick}}]
			\node {\pgfimage[width = 0.31\textwidth]{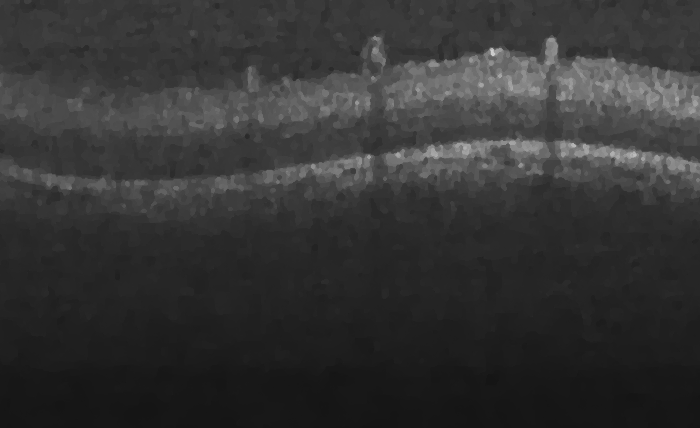}};
			\spy on (0.1,0.85) in node [left] at (1.9, -0.45);     
    \end{tikzpicture}}
	\caption{Comparison of simple averaging of consecutive B-scans (top row) to the proposed spatio-temporal denoising with TV regularization only (second row) and  TV + QuaSI regularization (third row) for different numbers of input images.}
	\label{fig:pigEyeImages}
\end{figure}
Fig.~\ref{fig:pigEyeImages} compares our method with and without the proposed QuaSI prior to simple averaging of consecutive B-scans for different numbers of input images. Notice that spatio-temporal denoising substantially enhanced noise reduction compared to pure averaging. We also found that regularization by our QuaSI prior in combination with the TV prior further enhanced noise reduction compared to pure TV denoising. Fig.~\ref{fig:pigEyeQualityMeasures} depicts the mean peak-signal-to-noise ratio (PSNR) and structural similarity (SSIM) relative to the gold standard. Here, our spatio-temporal method based on 4 B-scans only was comparable to averaging of 13 B-scans. Compared to pure TV denoising, the QuaSI prior enhanced the mean PSNR and SSIM by 0.9\,dB and 0.03, respectively. The proposed method also outperformed BM3D, BED and WMF in terms of both measures.
\\
\\
\textbf{Clinical Data.}
We also investigate denoising on clinical data which were acquired using a prototype ultrahigh-speed swept-source OCT system with 1050\,nm wavelength and a sampling rate of 400,000 A-scans per second \cite{Choi2013a}. Each B-scan was acquired five times in direct succession and the B-scans were registered towards the central one using cross-correlation. We use OCT data from 14 human subjects with two volumes each. The data covers proliferative and non-proliferative diabetic retinopathy, early age-related macular degeneration and one healthy subject. The field size is 3$\times$3\,mm with 500 A-scans by 500 B-scans. Our experiments were conducted on the central B-scan of each volume. To quantify noise reduction in the absence of a gold standard, we use the mean-to-standard-deviation ratio $\mathrm{MSR} = \mu_{f} / \sigma_{f}$ and the contrast-to-noise ratio $\mathrm{CNR} = \vert \mu_{f} - \mu_{b} \vert / (0.5(\sigma_{f}^{2} + \sigma_{b}^{2}))^{0.5}$ \cite{Fang2012,Ozcan2007,Wong2010}, where $\mu_{i}$ and $\sigma_{i}$, $i \in \{f, b\}$ denote the mean and standard deviation of the intensity in foreground ($i = f$) and background ($i = b$) image regions, respectively. Both measures were determined for five foreground regions and one background region that were manually selected for each B-scan, see Fig.~\ref{fig:patientDataImages:noisy}. 

\begin{figure}[!t]
	\scriptsize
	\centering
	\setlength \figurewidth{0.40\textwidth}
	\setlength \figureheight{0.60\figurewidth}
	\subfloat{\includegraphics[width = 0.60\textwidth]{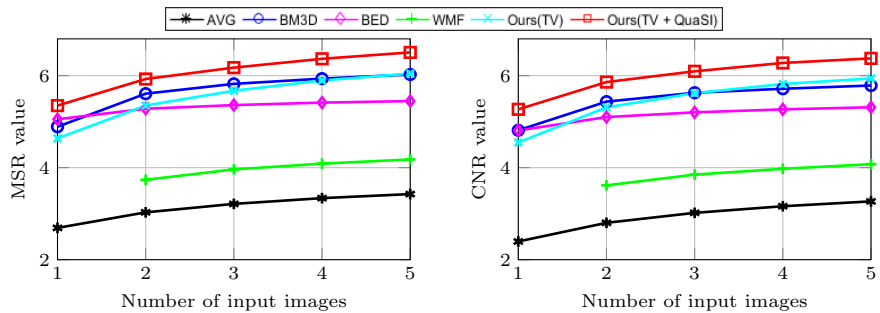}}\\[-1.0em]
	\subfloat{
%
%
\definecolor{mycolor1}{rgb}{1.00000,0.00000,1.00000}%
\definecolor{mycolor2}{rgb}{0.00000,1.00000,1.00000}%
\begin{tikzpicture}

\begin{axis}[%
width=0.951\figurewidth,
height=\figureheight,
at={(0\figurewidth,0\figureheight)},
scale only axis,
unbounded coords=jump,
xmin=1,
xmax=5,
xlabel style={font=\color{white!15!black}},
xlabel={Number of input images},
ymin=2.0,
ymax=6.8,
ylabel style={font=\color{white!15!black}},
ylabel={MSR value},
axis background/.style={fill=white},
xmajorgrids,
ymajorgrids,
ylabel near ticks,
xlabel near ticks
]
\addplot [color=black, line width=1.0pt, mark=asterisk, mark options={solid, black}, forget plot]
  table[row sep=crcr]{%
1	2.69106639550072\\
2	3.02872650259265\\
3	3.21284619974309\\
4	3.33781460188218\\
5	3.42592603315568\\
};
\addplot [color=blue, line width=1.0pt, mark=o, mark options={solid, blue}, forget plot]
  table[row sep=crcr]{%
1	4.89035099279165\\
2	5.60869098500747\\
3	5.82193047209191\\
4	5.93500052661609\\
5	6.02623563055597\\
};
\addplot [color=mycolor1, line width=1.0pt, mark=diamond, mark options={solid, mycolor1}, forget plot]
  table[row sep=crcr]{%
1	5.05290921429193\\
2	5.28105966580538\\
3	5.36022692987904\\
4	5.41457081240482\\
5	5.44765359161047\\
};
\addplot [color=green, line width=1.0pt, mark=+, mark options={solid, green}, forget plot]
  table[row sep=crcr]{%
1	nan\\
2	3.73656834032453\\
3	3.96248230412325\\
4	4.08782388667148\\
5	4.17884741284344\\
};
\addplot [color=mycolor2, line width=1.0pt, mark=x, mark options={solid, mycolor2}, forget plot]
  table[row sep=crcr]{%
1	4.63711939953742\\
2	5.35108874027202\\
3	5.67225843029814\\
4	5.89506116665507\\
5	6.04472413786319\\
};
\addplot [color=red, line width=1.0pt, mark=square, mark options={solid, red}, forget plot]
  table[row sep=crcr]{%
1	5.34960463670639\\
2	5.92770258160182\\
3	6.17504378486\\
4	6.36666123885143\\
5	6.50530049354036\\
};
\end{axis}
\end{tikzpicture}%
}\qquad
  \subfloat{
%
%
\definecolor{mycolor1}{rgb}{1.00000,0.00000,1.00000}%
\definecolor{mycolor2}{rgb}{0.00000,1.00000,1.00000}%
\begin{tikzpicture}

\begin{axis}[%
width=0.951\figurewidth,
height=\figureheight,
at={(0\figurewidth,0\figureheight)},
scale only axis,
unbounded coords=jump,
xmin=1,
xmax=5,
xlabel style={font=\color{white!15!black}},
xlabel={Number of input images},
ymin=2.0,
ymax=6.8,
ylabel style={font=\color{white!15!black}},
ylabel={CNR value},
axis background/.style={fill=white},
xmajorgrids,
ymajorgrids,
ylabel near ticks,
xlabel near ticks
]
\addplot [color=black, line width=1.0pt, mark=asterisk, mark options={solid, black}, forget plot]
  table[row sep=crcr]{%
1	2.39617760793673\\
2	2.80122991799017\\
3	3.01805162068995\\
4	3.16202229013265\\
5	3.26741060068456\\
};
\addplot [color=blue, line width=1.0pt, mark=o, mark options={solid, blue}, forget plot]
  table[row sep=crcr]{%
1	4.81291837009302\\
2	5.43546344751274\\
3	5.62862691241854\\
4	5.7155262651459\\
5	5.78622302664366\\
};
\addplot [color=mycolor1, line width=1.0pt, mark=diamond, mark options={solid, mycolor1}, forget plot]
  table[row sep=crcr]{%
1	4.80875446779945\\
2	5.09981102434438\\
3	5.20230044112208\\
4	5.26754526812077\\
5	5.31317203148205\\
};
\addplot [color=green, line width=1.0pt, mark=+, mark options={solid, green}, forget plot]
  table[row sep=crcr]{%
1	nan\\
2	3.61554910451747\\
3	3.84960971233251\\
4	3.9755332540374\\
5	4.07674970925513\\
};
\addplot [color=mycolor2, line width=1.0pt, mark=x, mark options={solid, mycolor2}, forget plot]
  table[row sep=crcr]{%
1	4.54848188452904\\
2	5.31081727385943\\
3	5.61980833289849\\
4	5.81892322773898\\
5	5.93921255349812\\
};
\addplot [color=red, line width=1.0pt, mark=square, mark options={solid, red}, forget plot]
  table[row sep=crcr]{%
1	5.26853913882987\\
2	5.86101623521823\\
3	6.09507140137894\\
4	6.27653773876636\\
5	6.37573957420352\\
};
\end{axis}
\end{tikzpicture}%
}
  \caption{Mean MSR and CNR measures to quantify noise reduction on our clinical dataset for different numbers of input images.}
  \label{fig:patientDataQualityMeasures}
\end{figure}

Fig.~\ref{fig:patientDataQualityMeasures} depicts the mean MSR and CNR for different numbers of input images. Here, BM3D and our spatio-temporal method achieved the best denoising performance in terms of both measures. In particular, the combination of the TV and QuaSI priors consistently outperformed the competing methods. Fig.~\ref{fig:patientDataImages} compares our approach and two competing methods on one example dataset. WMF enabled structure-preserving denoising but suffered from noise breakthroughs in homogeneous areas resulting in lower MSR and CNR measures. BM3D enables a strong noise reduction but suffered from streak artifacts as also noticed in other studies on OCT denoising \cite{Fang2012}, see the magnified image regions. The proposed method achieved a decent tradeoff between noise reduction and structure preservation. 

\section{Conclusion}
\label{sec:Conclusion}


This paper proposed a spatio-temporal denoising algorithm for OCT data. To effectively reduce speckle noise and to preserve morphological structures, we introduced the class of QuaSI priors for our energy minimization formation. We implemented this model via median filter regularization and devolved an ADMM scheme for efficient numerical optimization. Our method can handle denoising on the basis of single or multiple registered B-scans. Compared to simple B-scan averaging and state-of-the-art single-image methods, our algorithm is more effective in reducing speckle noise. In contrast to pure temporal methods, \eg \cite{Mayer2012}, we can adjust the number of B-scans as a tradeoff between denoising performance and acquisition time to the needs in specific applications. In our future work, we study the impact of our algorithm to common OCT image analysis tasks.

\bibliographystyle{splncs03}
\bibliography{bibliography}

\begin{thebibliography}{10}
\providecommand{\url}[1]{\texttt{#1}}
\providecommand{\urlprefix}{URL }

\bibitem{Cheng2014}
Cheng, J., Duan, L., Wong, D.W.K., Tao, D., Akiba, M., Liu, J.: {Speckle
  Reduction in Optical Coherence Tomography by Image Registration and Matrix
  Completion}. Proc. MICCAI 2014  8673,  162--169 (2014)

\bibitem{Choi2013a}
Choi, W., Potsaid, B., Jayaraman, V., Baumann, B., Grulkowski, I., Liu, J.J.,
  Lu, C.D., Cable, A.E., Huang, D., Duker, J.S., Fujimoto, J.G.:
  {Phase-sensitive swept-source optical coherence tomography imaging of the
  human retina with a vertical cavity surface-emitting laser light source}. Opt
  Lett  38(3),  338 (2013)

\bibitem{Dabov2007}
Dabov, K., Foi, A., Katkovnik, V., Egiazarian, K.: {Image Denoising by Sparse
  3-D Transform-Domain Collaborative Filtering}. IEEE Trans Image Process
  16(8),  145--149 (2007)

\bibitem{Duan2016}
Duan, J., Lu, W., Tench, C., Gottlob, I., Proudlock, F., Samani, N.N., Bai, L.:
  {Denoising optical coherence tomography using second order total generalized
  variation decomposition}. Biomed Signal Process Control  24,  120--127 (2016)

\bibitem{Fang2012}
Fang, L., Li, S., Nie, Q., Izatt, J.A., Toth, C.A., Farsiu, S.: Sparsity based
  denoising of spectral domain optical coherence tomography images. Biomed.
  Opt. Express  3(5),  927--942 (2012)

\bibitem{Goldstein2009}
Goldstein, T., Osher, S.: The split bregman method for l1-regularized problems.
  SIAM J Imaging Sci  2(2),  323--343 (2009)

\bibitem{Pan2016}
{Jinshan Pan, Deqing Sun, Hanspeter Pfister, Ming-Hsuan Yang}: Blind image
  deblurring using dark channel prior. Proc. CVPR 2016 pp. 1628--1636 (2016)

\bibitem{Kohler2015}
K\"{o}hler, T., Bock, R., Hornegger, J., Michelson, G.: {Computer-Aided
  Diagnostics and Pattern Recognition: Automated Glaucoma Detection}. In:
  Michelson, G. (ed.) Teleophthalmology in Preventive Medicine, pp. 93--104.
  Springer (2015)

\bibitem{Mayer2012}
Mayer, M.A., Borsdorf, A., Wagner, M., Hornegger, J., Mardin, C.Y., Tornow,
  R.P.: {Wavelet denoising of multiframe optical coherence tomography data}.
  Biomed Opt Express  3(3),  572 (2012)

\bibitem{Ochs2015}
Ochs, P., Dosovitskiy, A., Brox, T., Pock, T.: {On Iteratively Reweighted
  Algorithms for Nonsmooth Nonconvex Optimization in Computer Vision}. SIAM J
  Imaging Sci  8(1),  331--372 (2015)

\bibitem{Ozcan2007}
Ozcan, A., Bilenca, A., Desjardins, A.E., Bouma, B.E., Tearney, G.J.: {Speckle
  reduction in optical coherence tomography images using digital filtering}.
  Journal of the Optical Society of America A  24(7),  1901 (2007)

\bibitem{Pircher2003}
Pircher, M., Götzinger, E., Leitgeb, R., Fercher, A.F., Hitzenberger, C.K.:
  {Speckle reduction in optical coherence tomography by frequency compounding}.
  J Biomed Opt  8(3),  565 (2003)

\bibitem{Romano2016a}
Romano, Y., Elad, M., Milanfar, P.: {The Little Engine that Could:
  Regularization by Denoising (RED)}. arXiv preprint arXiv:1611.02862  (2016)

\bibitem{Salinas2007}
Salinas, H., Fernandez, D.: {Comparison of PDE-Based Nonlinear Diffusion
  Approaches for Image Enhancement and Denoising in Optical Coherence
  Tomography}. IEEE Trans Med Imaging  26(6),  761--771 (2007)

\bibitem{Wong2010}
Wong, A., Mishra, A., Bizheva, K., Clausi, D.a.: {General Bayesian estimation
  for speckle noise reduction in optical coherence tomography retinal imagery.}
  Opt Express  18(8),  8338--8352 (2010)

\end{thebibliography}

\end{document}